\documentclass{article}
\usepackage{textcomp,float}
\usepackage[numbers,sort&compress]{natbib}
\usepackage{tablefootnote}
\usepackage{booktabs}

\usepackage{amsmath,graphicx}
\usepackage[preprint]{spconf}

\title{Emotion recognition by fusing time synchronous and\\time asynchronous representations}
\name{Wen Wu, Chao Zhang, Philip C. Woodland\thanks{Wen Wu is funded by Cambridge Trust and China Scholarship Council.}}
\address{Cambridge University Engineering Dept., Trumpington St., Cambridge, CB2 1PZ U.K.\\
\small{\texttt{\{ww368,cz277,pcw\}@eng.cam.ac.uk}}}

\begin{document}

\copyrightnotice{\copyright\ IEEE 2021}

\ninept
\maketitle

\begin{abstract}
%Emotion recognition is an essential ability for machines to understand and interact with human. 
In this paper, a novel two-branch neural network model structure is proposed for multimodal emotion recognition, which consists of a time synchronous branch (TSB) and a time asynchronous branch (TAB). 
%, which uses a time synchronous branch to fuse speech with text along with a time asynchronous branch to 
To capture correlations between each word and its acoustic realisation, the TSB combines speech and text modalities at each input window frame and then uses pooling across time to form a single embedding vector. 
%with a  multi-head self-attentive layer.
%TAB, on the other hand, provides cross-utterance information by integrating multiple BERT embeddings for the texts from a number of context utterances into another embedding vector.
The TAB, by contrast, provides cross-utterance information by integrating sentence text embeddings from a number of context utterances into another embedding vector. 
The final emotion classification uses both the TSB and the TAB embeddings. Experimental results on the IEMOCAP dataset demonstrate that the two-branch structure achieves state-of-the-art results in 4-way classification with all common test setups.
%in the four-way classification setup tested with session 5, 5-fold cross-validation (CV) and 10-fold CV. 
When using automatic speech recognition (ASR) output instead of manually transcribed reference text, it is shown that the cross-utterance information considerably improves robustness against ASR
 errors. 
%Further by incorporating an extra class for ``frustrated'' and all the other emotions, our final 5-way classifcation system with ASR outputs can be a prototype of more realistic multimodal emotion recognition systems. 
Furthermore, by incorporating an extra class for all the other emotions, the final 5-way classification system with ASR hypotheses can be viewed as a prototype for 
more realistic emotion recognition systems.

\end{abstract}
%
%\begin{keywords}
%Emotion recognition, multimodal
%\end{keywords}
%
\section{Introduction}
\label{sec:intro}
% One of the major tasks of intelligent human-machine interaction is to empower computers with the ability of ``affective computing" such that it can recognize a user’s emotional status and respond to the user in an affective way. 
%Automatic emotion recognition (AER) recognition, as an essential ability for machines to understand and interact with human, has attracted much attention due to its wide range of potential applications in, for example, driver monitoring, mental health analysis, spoken dialogue system and chatbot. Although significant progress has been made \cite{Kim_2013, Poria_2018}, AER is still a challenging research problem since human emotions are inherently complex, ambiguous, and highly personal, and the training and test data are often insufficient due to the difficulties and considerations in data collection and protection. Current research efforts include the use of transfer learning with speech or speaker recognition data \cite{Latif2018TransferLF,Lu_ASR_2020}, multi-task learning with gender or speaker classification to model the personal bias of emotions \cite{Nediyanchath_2020}, features embedded in multiple modalities \cite{Poria_2018, Yoon_2019}, and more powerful model architectures \cite{Majumder_2018,HGFM}.
Automatic emotion recognition (AER) is an essential capability for machines to understand and interact with humans and has attracted much attention due to its wide range of potential applications in {\it e.g.} driver monitoring, mental health analysis, spoken dialogue systems and chatbots. Although significant progress has been made \cite{Kim_2013,Poria2017,Tzirakis2017}, AER is still a challenging research problem since human emotions are inherently complex, ambiguous, and highly personal. Humans often express their emotions  using multiple simultaneous approaches, such as voice characteristics, linguistic content, facial expressions, and body actions, which makes AER by nature a complex multimodal task \cite{Pantic_2005,Soleymani_2012,Mittal_2020}. Furthermore, due to the difficulties in data collection, publicly available datasets often do not have enough speakers to properly cover personal variations in  emotion expression. 
Consequently, current research efforts include the use of transfer learning with speech or speaker recognition data \cite{Latif2018TransferLF,Lu_ASR_2020}, multi-task learning with gender or speaker classification to model the personal aspects of emotions \cite{Nediyanchath_2020},  features embedded in multiple modalities, and more powerful model architectures. For instance, various types of acoustic features can be fused with text features derived either from pre-trained word embeddings \cite{tripathi2018multimodal,Yoon_2018} or from a jointly trained neural network component \cite{Poria_2018,Yoon_2019}. Context-dependent hierarchical fusion \cite{Majumder_2018,HGFM}, multi-head attention mechanisms \cite{Yoon_2019}, and multiplicative fusion \cite{Mittal_2020} have been applied to emotion recognition.

In this paper, we propose a novel deep neural network architecture for AER, which consists of a time synchronous branch (TSB) and a time asynchronous branch (TAB). In the TSB, audio features are fused with the corresponding text at each frame in the input window. This can model not only the speech content and its acoustic realisation, but also their alignments and can help capture prosody. Note that video features could also be optionally included in the TSB.
Along with traditional acoustic features, we propose using long-term acoustic features with a 250 millisecond (ms) frame length as an extra type of audio feature. In contrast to the TSB that focuses on modelling the correlations of multimodal features across time, the TAB leverages the transcriptions of the spoken utterances as well as other potential global or utterance-level features.  
%to focus on capturing the semantic meanings of the speech. 
Sentence embeddings relevant to a number of consecutive utterances in the dialogue are 
derived from a pre-trained model with bidirectional encoder representations from Transformers (BERT) \cite{devlin-etal-2019-bert}, and are used as the cross-utterance input to the TAB. Two self-attentive layers \cite{Lin2017ASS}  are used as the separate pooling functions for the TSB and TAB, and the resulting vectors fused for final emotion classification. Experimental results on the widely used IEMOCAP dataset \cite{Busso2008IEMOCAPIE} show that the proposed structure achieves state-of-the-art results in the usual 4-way classification setup when evaluated with all usual test configurations. Rather than assuming perfect automatic speech recognition (ASR), the use of ASR outputs was also studied, and it was found that using cross-utterance input considerably improves robustness against ASR errors. 
Finally, a 5-way classification setup is studied, which has an extra class to represent ``frustration'' and all the other emotions in IEMOCAP that are not considered in the 4-way setup.

The rest of the paper is organised as follows. Section~\ref{sec:method} introduces our proposed two-branch structure. Section~\ref{sec:exp} presents the experimental setup and results related to the model architecture. Further analysis and experiments towards implementing a more realistic AER system is given in Section~\ref{sec:discussion}, followed by conclusions.
%We conclude in Section~\ref{sec:conclusion}.

\section{Proposed approach}
\label{sec:method}

\subsection{Feature representation}
\subsubsection{Audio features}

%Mel-Frequency Log Filerbank Energies (MFBs) and pitch are used as audio features. MFBs have been shown to be better discriminative features than MFCCs in emotion recognition~\cite{Busso_2007}. This paper uses 40-channel log energies extracted using a 25ms window with a frame shift of 10ms (MFB$_{25}$).

%Filterbank information describes vocal tract shape and vocal spectrum but doesn't include information about excitation. Pitch can provide complementary information. It has been suggested that acoustic parameters such as pitch plays a crucial role in obtaining better performance in speech emotion recognition (SER)~\cite{Pappagari_2020}. Many studies has established a link on prosody attributes that high pitch levels are related to emotions carrying a high level of activity, such as joy, anxiety, or fear; medium pitch levels account for more neutral attitudes; low pitch levels are related to sober emotions: sadness, calmness, or security~\cite{pitch_emotion}. In this paper, log-pitch with Probability of Voicing (POV)-weighted mean subtraction over a 1.5s window are used.
%Early research shows that long term audio features can help improve the performance of SER systems~\cite{Dumouchel2009CepstralAL}. Although some of the long-term audio features can be implicitly captured with the help of neural networks, it is worth introducing some explicitly. The use of log energies extracted using 250ms window with a frame advance of 10ms (MFB$_{250}$) as an additional audio feature is investigated later in this paper.

The audio representation for speech-based AER often includes log Mel filterbank features (FBKs) \cite{Busso_2007}. In this paper, 40-dimensional (-d) FBKs with a 10~ms frame duration and 25 ms frame length are used, which is denoted FBK$_{25}$. FBK features have information about the short-term spectrum but do not contain pitch information that can be important in describing emotional speech \cite{pitch_emotion} and is often complementary to FBKs \cite{Dumouchel2009CepstralAL,Pappagari_2020}. The log pitch frequency features with probability-of-voicing-weighted mean subtraction over a 1.5 second window are used along with FBKs \cite{pitch2014}. In addition, we propose using long-term FBK features extracted in the same way as FBK$_{25}$ apart from a long 250ms frame length, which is denoted as FBK$_{250}$.

\subsubsection{Text embeddings}
The use of vector representations of words and sentences has become a widely-used approach in natural language processing. The global vector (GloVe) is a commonly used method that estimates word embeddings using a global log-bilinear regression model in order to combine
the advantages of both global matrix
factorization and local context window
methods \cite{Glove}. In this paper, pre-trained 50-d GloVe embeddings are used in the TSB to encode word-level transcriptions. Recently, pre-trained sentence-level embeddings derived from a  large pre-trained language model, such as BERT \cite{devlin-etal-2019-bert}, have drawn much attention. %, due to its excellent performance across various natural language understanding tasks. 
We use the pre-trained BERT-base model without fine-tuning to encode the transcription of each single utterance into a 768-d vector, which is used as the input to the TAB.  

%50-d GloVe vectors pre-trained on Twitter were used in this section. The Twitter database contains 2B tweets, 27B tokens, and a 1.2M vocabulary. 

\subsection{Model structure}
\label{sec:tsbtab}
\begin{figure}[htbp!] 
\centering    
\includegraphics[width=\linewidth]{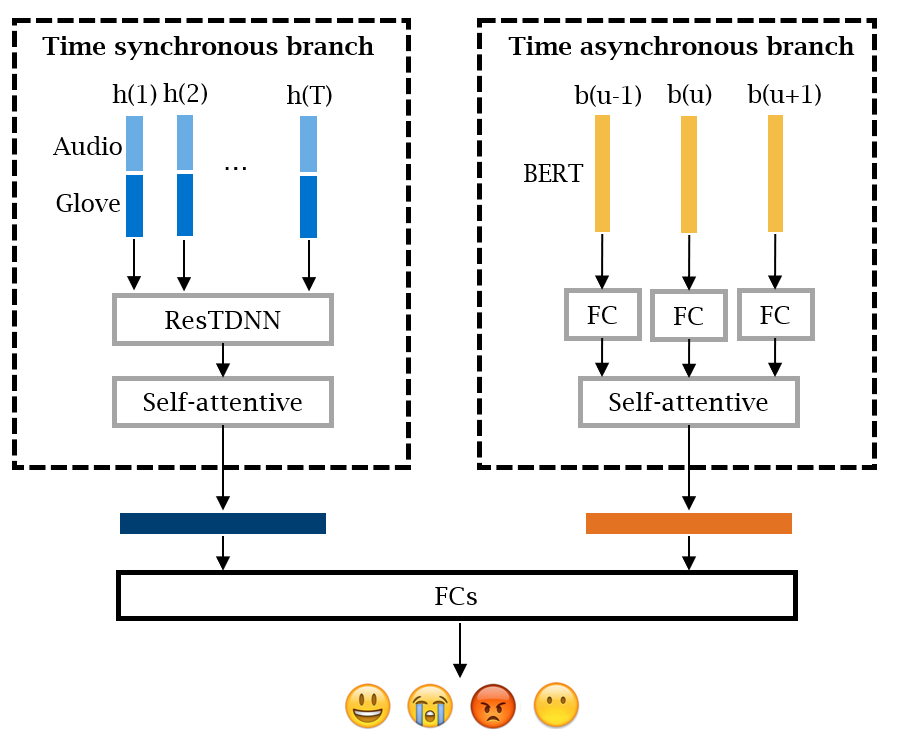}
\vspace{-7mm}
\caption{Proposed two-branch model structure.}
\vspace{-2mm}
\label{fig:final}
\end{figure}

The proposed model structure is shown in Fig.~\ref{fig:final}, which consists of a TSB that fuses the audio features with the  corresponding text information at each time step, as well as a TAB that captures the  text information embedded across the transcriptions of a number of consecutive utterances.
The TSB structure is similar to that which is often used for speaker embedding extraction~\cite{Sun_2019}. The TSB uses a five-head self-attentive layer \cite{Lin2017ASS} to pool the frame-level vectors across time in the input window, and a time delay neural network with residual connections \cite{Kreyssig2018ImprovedTU} is used as the encoder to derive the frame-level vectors. A modified penalty term is used for the five-head self-attentive layer, with three heads set to obtain more ``spiky'' attention weight distributions and the other two set to obtain  ``smoother'' distributions \cite{Sun_2019}. 
%has been adopted and adapted for TSB: time delay neural networks with residue connection (ResTDNN)~\cite{Kreyssig2018ImprovedTU} acting as encoder and five-head self-attentive layers \cite{Lin2017ASS} with the modified penalty term \cite{Sun_2019} performing temporal pooling. 
In the TSB, the audio features and the corresponding GloVe-based word embeddings are combined at each time step with a simple concatenation operation. This structure exploits the alignment between each speech unit and its acoustic realisation. 
% to capture their correlations. 
This could be particularly useful for modelling prosody, including the duration for each speech unit and pitch information, which is important for emotional expressions in spoken language \cite{10.3389/fnins.2016.00506}. Moreover, video information can also be included in the TSB, by simply concatenating the visual features of each video frame with the corresponding audio and text features, and ensuring that the frame rate of the audio features matches the video frame rate. Since we did not observe any performance improvement by including the visual features, possibly due to the quality and style of the videos in IEMOCAP, no video-related experiments are presented in this paper. More detail of these video-related experiments can be found in \cite{ww2020thesis}. 
 
%Apart from the speech content and its acoustic realisation, this fusion method also provides alignment information between them, which can help the model better exploit the correlations between two modalities. For example, as one word usually lasts for multiple frames, each word embedding may be repeated for several time steps. This repetition provides word duration information and is important clue to predict emotional content. As emotional labels are often provided for the whole utterances in many databases including IEMOCAP, there can be a mismatch between short-term inputs at frame level (i.e., the 10ms acoustic features) and long-term outputs at the utterance level. The use of attention pooling resolves this mismatch, resulting in one embedding vector per input window. 

% context BERT 

While the TSB includes modelling the temporal correlations between different modalities, the TAB focuses on capturing text information including meaning from the speech transcriptions. The BERT-derived sentence embeddings of the utterance transcriptions are used as the input vectors to TAB. The embeddings for a number of consecutive utterances were used as the TAB input since the emotion of each utterance is often strongly related to its context in a spoken dialogue~\cite{Mark_2012}.  %For example, context sentences have been used for sentiment clustering in Text To Speech (TTS)~\cite{Mark_2012}.
%rather than only using the BERT embedding for the current utterance, 
The following text snippets illustrate the importance of context in determining emotional content:

\begin{table}[H]
\vspace{-2mm}
\centering
\begin{tabular}{l}
``Did you pass the exam?'' \\
``Yes, exactly.'' (happy)\\
\hline
``Did you fail the exam?''\\
``Yes, exactly.'' (frustrated)
\end{tabular}
\vspace{-3mm}
\end{table}
\noindent It is clear that the two different examples of ``Yes, exactly.'' convey very different emotions, and hence determining the emotion from the text requires the context. Furthermore, since cross-utterance information provides extra information for the emotional content, it is possible to improve the AER system robustness when using erroneous transcriptions derived from an ASR system. 
In the TAB, a shared fully-connected (FC) layer is used to reduce the dimension of each input BERT embedding, and the resulting vectors are then integrated by  
 %the dimension of each BERT embedding is reduced by an FC layer. The weights and biases are shared among these FCs. The low-dimensional context BERT embeddings are integrated by 
another five-head self-attentive layer, whose attention-weight distribution reflects the extent to which sentences in the context affect the current emotion.

 %However, including sentence embedding of consecutive sentences in TSB can be computationally expensive. An addition TSB is then introduced to incorporate cross-utterance information. It integrates multiple sentence embeddings extracted from language models, such as BERT, into one embedding vector.
% Contextual information are crucial to accurately classify emotions.

Finally, the output vectors from both branches are fused using an FC layer for emotion classification. The hidden and output activation functions are ReLU and softmax respectively, and a large-margin softmax loss function is used to avoid over-fitting \cite{SphereFace}.

%All hidden layers use ReLU function except for the output layer with a angular softmax~\cite{SphereFace}. 

\section{Experiments}
\label{sec:exp}

%\subsection{Dataset}
\subsection{Experiment setup}

The IEMOCAP~\cite{Busso2008IEMOCAPIE} corpus used in this paper is a multimodal dyadic conversational dataset. It consists of approximately 12 hours of multimodal data, including speech, text transcriptions and facial recordings. IEMOCAP contains a total of 5 sessions and 10 different speakers, with a session being a conversation of two exclusive speakers. To be consistent and to be able to compare with previous studies, only utterances with ground truth labels 
%\footnote{Defined as having majority vote agreement among human labellers.}  
belonging to ``angry", ``happy", ``excited", ``sad", and ``neutral" were used. The ``excited'' class was merged with ``happy'' to better balance the size of each emotion class, which results in a total of 5,531 utterances (happy 1,636, angry 1,103, sad 1,084, neutral 1,708).

%\subsection{Experiment setup}

Unless otherwise stated, leave-one-session-out 5-fold cross validation (CV) is used and the average result reported. At each fold of the 5-fold CV set-up, 8 speakers are used for training while the other two are used for testing. Since the test sets are slightly imbalanced between different emotion categories, both the weighted accuracy (WA) and unweighted accuracy (UA) are reported. Models were implemented using HTK~\cite{HTK} in combination with PyTorch. The newbob learning rate scheduler with an initial learning rate of $5\times 10^{-5}$ was used throughout training.

\subsection{TSB only results}

Dialogue-level variance normalization and utterance-level mean normalization were performed on audio features. Pitch and the first differentials were appended to the 40-d FBK$_{25}$. The released reference transcripts of IEMOCAP were used for the text modality as most previous work on IEMOCAP takes this approach. 
Although \cite{tripathi2018multimodal,Yoon_2018,Yoon_2019} used 300-d GloVe in their experiments, 50-d GloVe was found to be most effective in our framework.
As shown in Table~\ref{tab:A+T}, attaching GloVe to the audio features improves the accuracy by $\sim$5\% absolute. The standard deviation across the five folds also decreases when these two features are combined, which indicates that the system is more robust to speaker variation.

%The 5-cv results of the system using different input audio features are shown in Table~\ref{tab:audio}. Comparing rows 2 and 3, pitch information is useful. It improves the results by \texttildelow 0.75\% by only adding one more dimension. Although it can be argued that dynamic features can be redundant if they are combined with a linear layer in a subsequent neural network, appending the first differentials still improves the results by \texttildelow 2.2\%. The standard deviation also decreases when deltas are appended.\\

\iffalse
\begin{table}[htbp!]
\centering
\begin{tabular}{cccc}
\hline
Feature         & dim &  WA (\%) & UA (\%) \\
\hline
FBK             & 40     & 57.20$\pm$1.84  & 58.78$\pm$3.21  \\
FBK+pitch       & 41     & 57.91$\pm$2.17  & 59.59$\pm$3.40  \\
FBK+pitch+$\Delta$ & 82  & 60.64$\pm$1.96 & 61.32$\pm$2.26\\
\hline
\end{tabular}

\caption{Average and standard deviation of 5-fold CV results on audio-based AER system using different combination of audio features as input.}
\label{tab:audio}
\end{table}
\fi

%GloVe vectors are attached to audio features at the frame level to form a feature-level concatenation fusion. 

%The input audio features have a context range from -2 to +2. GloVe is only attached to the central frame as one word lasts for several frames and the neighbouring frames are very likely to be the same.

\begin{table}[htbp!]
\centering
\begin{tabular}{cccc}
\toprule
Feature  & WA (\%) & UA (\%) \\
\midrule
Audio$_{25}$  & 60.64$\pm$1.96 & 61.32$\pm$2.26\\
GloVe   & 61.27$\pm$3.73     &62.67$\pm$3.55  \\
Audio$_{25}$+GloVe   & 65.53$\pm$1.83  & 66.43$\pm$1.33 \\
\bottomrule
\end{tabular}
\caption{Mean and std. dev. of the TSB only 5-fold CV results with different input features. ``Audio$_{25}$" denotes FBK$_{25}$+pitch+$\Delta$.}
\vspace{-5mm}
\label{tab:A+T}
\end{table}

%\vspace{-2mm}
\subsection{Results with TAB}
\label{sec: bert}
Here we present the results with both the TSB and TAB branches. 
%BERT embeddings of consecutive utterances were extracted from a pretrained model \cite{Wolf2019HuggingFacesTS}. 
%IEMOCAP is a dyadic database. The context can be chosen to either only include context of the current speaker or include context of both speakers in dialogue turn.
Table~\ref{tab: context bert} shows the results vary with the  number of context utterances from the dialogue. Comparing systems with no context ([0]) to 4 utterances 
before and 4 after ([-4,4]), shows that increased context leads to improved accuracy. The best UA and WA results were obtained with context windows of [-3,3] and [-4,4] respectively. 
%while Row num.4 and 5 give comparable results. Comparing Row num.3,6 and 4,7, with the same context width, including both speakers in context performs better. 
%It is arguable that emotion is a long-term attribute that may last for several utterances, and the use of context utterances from the same speaker helps to validate the current emotion. As in a dialogue, cross-utterance information from the other speaker carries reaction information, which can also be useful to infer the current speaker's emotion. What the other speaker says may affect the listener's emotion to a considerable extent and can trigger a change in the listener's emotion. 
Since emotion is a long-term attribute that may last for several utterances, context utterances from the same speaker help to recognise the current emotion. As a dyadic dialogue, utterances spoken by the other speaker carry  reaction information, which is also useful to infer the current speaker's emotion.  
The systems with [-2,0] and [-3,0] contexts do not use information from future utterances and simulate an online streaming AER system and demonstrates the importance of the future utterance information. A [-3,+3] context window is used for all future systems thereafter, unless otherwise stated. 

%The last two rows consider only previous sentences (in online learning manner). Comparing Row num.2 and Row num.8, using the same number of context utterances, the following sentence is more informative than that before the previous sentence. Comparison of Row num.3,8 and Row num.4,9 shows the effectiveness of including subsequent sentences. Context of [-3,+3] was selected for later experiments.
\begin{table}[htbp!]
\centering
\setlength{\tabcolsep}{1.5mm}
\begin{tabular}{cccccccc}
\toprule
Context & [0] & [-1,1] & [-2,2] & [-3,3] & [-4,4] & [-2,0] &  [-3,0] \\
\hline
UA (\%) & 70.83 & 77.60 & 77.76 & \textbf{81.22} & 80.66 & 74.94 & 78.81\\
WA (\%) & 71.61 & 77.90 & 79.20 & 81.60 & \textbf{81.87} &75.74 & 78.38\\
\bottomrule
\end{tabular}
\caption{Results of the two-branch systems with different TAB contexts, which were trained on Session 1--4 and tested on Session 5. }%\# speaker equals ``one" means that only sentences of the current speaker were included in context. ``Two" means that context of both speakers were included. Bold: highest value in each column. }
\label{tab: context bert}
\end{table}

%\begin{table}[htbp!]
%\centering
%\begin{tabular}{ccccccc}
%\hline
%Row num.& context       & \# speaker &  WA (\%)         & UA (\%)         \\
% \hline
%1 & {[}0{]}       & two           & 70.83          & 71.61          \\
%2 & {[}-1,+1{]}   & two           & 77.60          & 77.90          \\
%3 & {[}-2,+2{]}   & two           & 77.76          & 79.20          \\
%4 & {[}-3,+3{]}   & two           & \textbf{81.22} & 81.60          \\
%5 & {[}-4,+4{]}   & two          & 80.66          & \textbf{81.87}\\
% \hline
%6  & {[}-2,+2{]}   & one    & 75.02          & 76.71 \\
%7 & {[}-3,+3{]}   & one          & 80.66          & 80.99          \\
% \hline
%8   & {[}-2,0{]} & two           & 74.94          & 75.74          \\
%9   & {[}-3,0{]} & two           & 78.81          & 78.38          \\
%\hline
%\end{tabular}
%\caption{Results of the TSB-TAB system using different context configurations in TAB. Trained on Session1-4 and tested on Session 5. \# speaker equals ``one" means that only sentences of the current speaker were included in context. ``Two" means that context of both speakers were included. Bold: highest value in each column. }
%\label{tab: context bert}
%\end{table}
Table~\ref{tab:A+T+context_bert} shows the results of the TAB only system with no context utterances (BERT[0]) and a [-3,3]  context window (BERT[-3,3]), along with the two-branch system with a [-3,3] context window (Audio$_{25}$+GloVe+BERT[-3,3]).
%using single-utterance TAB alone (``BERT0"), cross-utterance TAB alone (``BERT7"), and the combination of TSB and cross-utterance TAB. 
It can be seen that the TAB only system with [-3,3] produces both a higher accuracy and a lower standard deviation. %It has much higher average accuracy as well as lower standard deviation. 
Compared to only using the  TSB  (Audio$_{25}$+GloVe in Table~\ref{tab:A+T}), incorporating cross-utterance information from the TAB improves the accuracy by  $\sim$10\%. %In the following content, unless otherwise stated, ``BERT" refers to ``BERT7".

\begin{table}[htbp!]
\centering
\begin{tabular}{cccc}
\toprule
Feature  & WA (\%) & UA (\%) \\
\midrule
BERT[0] & 58.53$\pm$4.41 & 59.20$\pm$5.57\\
BERT[-3,3] & 71.22$\pm$3.16  & 71.88$\pm$2.62\\
Audio$_{25}$+GloVe+BERT[-3,3]    & 75.53$\pm$3.79  & 76.65$\pm$3.67 \\
\bottomrule
\end{tabular}
\caption{Mean and std. deviation of the 5-fold CV results with different input features. ``Audio$_{25}$" denotes FBK$_{25}$+pitch+$\Delta$.} 
%using single-utterance TAB alone (``BERT0"), cross-utterance TAB alone (``BERT7"), and the combination of TSB and cross-utterance TAB.}
\label{tab:A+T+context_bert}
\end{table}
\vspace{-2mm}

\subsection{Ablation study with different features}
%\subsection{Contribution of different features and correlation between modalities}
%\begin{table}[tbp!]
%\centering
%\begin{tabular}{lccccc}
%\toprule
% Feature & Audio & GloVe & BERT & WA (\%) & UA (\%) \\
%\midrule
%Audio  & $\surd$ &       &      & 60.64  & 61.32  \\
%GloVe   &       & $\surd$ &      & 61.27  & 62.67  \\
%BERT    &       &       & $\surd$ & 71.22  & 71.88  \\
%\midrule
%Text   &       & $\surd$ & $\surd$ & 70.40   & 71.88  \\
%Audio+GloVe & $\surd$ & $\surd$ &   & 65.53  & 66.43  \\
%Audio+BERT  & $\surd$ &      & $\surd$ & 74.10  & 74.89  \\
%\midrule
%Audio+Text  &$\surd$ & $\surd$ &$\surd$ & 75.53  & 76.65 \\
%\bottomrule
%\end{tabular}
%\caption{5-CV mean values of systems with different features (\& modalities). Reference transcriptions are used for all text features. }
%\label{tab: ablation}
%\end{table}
\begin{table}[tbp!]
\centering
\begin{tabular}{cccccc}
\toprule
Audio$_{25}$ & GloVe & BERT & FBK$_{250}$ & WA (\%) & UA (\%) \\
\midrule
$\surd$ &       &      & & 60.64  & 61.32  \\
     & $\surd$ &      & & 61.27  & 62.67  \\
     &       & $\surd$ & & 71.22  & 71.88  \\
\midrule
     & $\surd$ & $\surd$ & & 70.40   & 71.88  \\
$\surd$ & $\surd$ &   & & 65.53  & 66.43  \\
$\surd$ &      & $\surd$ & & 74.10  & 74.89  \\
\midrule
$\surd$ & $\surd$ &$\surd$ & & 75.53  & 76.65 \\
$\surd$ &  & $\surd$ & $\surd$ & 74.74  & 75.60 \\
$\surd$ & $\surd$ & $\surd$ & $\surd$ & \textbf{76.12}  & \textbf{77.36} \\
\bottomrule
\end{tabular}
\caption{5-fold CV mean values of systems with different features. Reference transcriptions are used for all text features.} %``Audio$_{25}$" denotes FBK$_{25}$+pitch+$\Delta$pitch.}
\label{tab: ablation}
\vspace{-4mm}
\end{table}
The results of an ablation study are shown in Table~\ref{tab: ablation}. Although BERT is more effective than GloVe, the addition of GloVe can be useful. Comparing BERT-only, GloVe+BERT, Audio$_{25}$+BERT, and Audio$_{25}$+GloVe+BERT systems, adding GloVe to the BERT-only system can reduce the accuracy while adding GloVe to Audio$_{25}$+BERT system can increase accuracy, which shows the importance of the prosody provided through correlations between audio and GloVe.
Furthermore, by comparing the Audio$_{25}$+BERT and GloVe+BERT systems, it can be seen that although GloVe itself is more useful than audio features, audio features perform better when combined with BERT, which indicates that the audio features provide complementary information to the BERT embeddings.
%Furthermore, by comparing the BERT-only, audio+GloVe, and audio+BERT systems, it can be seen that BERT in TAB is more useful than GloVe in TSB, and the audio features are  complementary to the BERT embeddings.
%Comparing Audio+GloVe (Row num.5) and Audio+BERT (Row num.6), GloVe alone system (Row num.2) gives lower results than BERT only system (Row num.3) and Audio+GloVe also produces lower results than Audio+BERT system, as summarized in Equation~\ref{eqn: expected}.
%\begin{equation}
%\begin{gathered}
%    \text{GloVe} < \text{BERT}\\
%    \text{GloVe}+\text{Audio}<\text{BERT}+\text{Audio}
    %\text{bert+audio+fbk250}&<\text{bert+glove+audio+fbk250}
%\label{eqn: expected}
%\end{gathered}
%\end{equation}

%However, comparing GloVe+BERT (Row num.4) and Audio+BERT (Row num.6), although GloVe alone system (Row num.2) behaves better than audio alone system (Row num.1), Audio+BERT produces higher accuracies than GloVe+BERT, as summarized in Equation~\ref{eqn: compliment}. This may due to the fact that audio features provide more complimentary information to BERT than GloVe vectors do.

%\begin{equation}
%\begin{gathered}
%    \text{GloVe} > \text{Audio}\\
%    \text{GloVe}+\text{BERT}<\text{Audio}+\text{BERT}\\
%    %\text{bert+audio+fbk250}&<\text{bert+glove+audio+fbk250}
%\label{eqn: compliment}
%\end{gathered}
%\end{equation}

%\subsection{Long-term audio features}

Next, we study the use of our proposed long-term acoustic feature FBK$_{250}$. With added FBK$_{250}$ features, there is a $\sim$0.65\% increase in the classification accuracy. 
The best results achieve 76.12\% WA and 77.36\% UA for 5-fold CV. As shown in Table~\ref{tab: ablation}, GloVe is still useful even if the long-term acoustic features are used. 
%Since more features might require more regularization. Dropout with 0.5 dropout probability was added to the output of each residue block and each self-attentive layer. Adding dropout increases the overall performance by \texttildelow 1\%, resulting in 5-fold CV average of 77.57\% WA and 78.41\% UA.
Here we also found the use of long-term acoustic features requires more regularization. Dropout with a dropout rate of 0.5 was added to the output of each residual block and each self-attentive layer. Adding the dropout increases the overall performance by $\sim$1\%, resulting in the 5-fold CV averages of 77.57\% WA and 78.41\% UA.

\section{Discussion}
\label{sec:discussion}

\subsection{Cross comparisons}

%Time sync and Time async branch result

To compare to the best previous results published on IEMOCAP, both a single fold test with the model trained on Session 1-4 and tested on Session 5, as well as the leave-one-speaker-out test with 10-fold CV are provided for the final system, as shown in Table~\ref{tab: liter sum}. Our systems only used emotional audio data from IEMOCAP and two pre-trained text embeddings, without any audio data augmentation. The results and modalities used in the related work are summarised in Table~\ref{tab: liter sum}. %Although there are some other good results \cite{Mittal_2020}, they didn't state explicitly which test setting they used. 
%There is one other study \cite{Mittal_2020}, which although presented very good results, was not clear about what was the testing setting used. 
It is worth noting that although \cite{ijcai2019-703} produces higher 10-CV UA, their cross validation setup is not speaker exclusive.
% To the best of our knowledge, our final system gives the best WA and UA  results on IEMOCAP when evaluated with all of the three test settings. 

%\begin{table}[htbp!]
%\centering
%\begin{tabular}{ccc}
%\toprule
%Test & WA (\%) & UA (\%) \\
%\midrule
%Session 5 & 83.08  & 83.22 \\
%5-fold CV & 77.57  & 78.41 \\
%10-fold CV & 77.76 & 78.30\\
%\bottomrule
%\end{tabular}
%\caption{Emotion classification results using different training and testing setting. Averages across folds reported.} %``Ses05" denotes training on Session 1-4 and testing on Session 5. ``5-CV" denotes 5-fold leave-one-session-out cross validation. ``10-CV" denotes 10-fold leave-one-speaker-out cross validation. Averages across folds reported.}
%\label{tab: final system-10}
%\end{table}

\begin{table}[htbp!]
    \centering
    \scalebox{0.95}{
    \begin{tabular}{lcccc}
    \toprule
        Paper & Modality  & Test Setting & WA (\%) & UA (\%)\\
        \midrule
        ours & A+T & Session 5 & \textbf{83.08} & \textbf{83.22} \\
        ours & A+T & 5-fold CV & \textbf{77.57} & \textbf{78.41} \\
        ours & A+T & 10-fold CV & \textbf{77.76} & \textbf{78.30} \\
        \midrule
        %\cite{tripathi2018multimodal}& A+T+V & Session 5 & 71.04 & -- \\
        
        \cite{Majumder_2018} & A+T+V &  Session 5 & 76.5 & --\\

        \cite{Yoon_2018}&A+T & 5-fold CV & -- & 71.8 \\
        
        %\cite{Kim_2013}&A+V & 10-fold CV & -- & 66.12 \\

        \cite{Poria_2018}& A+T+V & 10-fold CV & 76.1 & --\\

        \cite{Yoon_2019}&A+T &10-fold CV & 76.5 & 77.6 \\
        
        \cite{ijcai2019-703} & A+T & 10-fold CV & -- & 79.2 \\

        %\cite{Yoon_2019}&A+T &10-fold CV\tablefootnote{Instead of the leave-one-speaker-out 10-fold CV, Yoon et al. used each 8, 1, 1 folds for the train set, development set, and test set, respectively.} & 76.5 & 77.6 \\
        
        %\cite{ijcai2019-703} & A+T & 10-fold CV\tablefootnote{Instead of the leave-one-speaker-out 10-fold CV, Li et al. randomly shifted IEMOCAP and divided it into three partitions with a proportion of 8:1:1 for training, validation and testing, which is not speaker exclusive.} & -- & 79.2 \\
        \bottomrule
    \end{tabular}}
    \caption{Summary of results in the literature. ``A'', ``T'', and ``V'' refer to the audio, text, and video modalities respectively. Results from \cite{Mittal_2020} are not included since the test setting was not clear. Results from \cite{Yoon_2019} and \cite{ijcai2019-703} were not obtained using leave-one-speaker-out 10-fold CV and thus not directly comparable. }
    \label{tab: liter sum}
    \vspace{-4mm}
\end{table}

\subsection{Use of ASR transcriptions}
Although the released reference transcriptions of the IEMOCAP dataset were used in all of the previous experiments, in practice, reference transcriptions are usually not available. Therefore, here we study the use of the transcriptions generated by a real ASR system. The Google Cloud Speech API
%\footnote{https://cloud.google.com/speech-to-text/} 
was used to generate the ASR transcriptions. Among all 5,531 utterances, 9.7\% did not receive any valid recognition output. The ASR system gives an overall WER of 40.8\%. This indicates that emotional speech recognition is still a difficult task for a standard ASR system. %This is similar to the findings reported in \cite{Sahu2019MultiModalLF}.

%In addition to the reference transcripts that most related work used for text modality, the use of transcriptions generated by ASR was also studied. Recognizing emotional speech is difficult. The ASR transcription which had WER above 40\% led to \texttildelow 10\% decrease in results of single BERT system. This decrease was reduced when context BERT and audio were included.
\begin{table}[htbp!]
\centering
\begin{tabular}{cccc}
\toprule
Feature     & Text & WA (\%) & UA (\%)\\
\midrule
BERT[0]       & Ref  & 58.53 & 59.20  \\
BERT[-3,3]       & Ref  & 71.22  & 71.88  \\
Audio$_{25,250}$+BERT[-3,3]  & Ref  & 74.74  & 75.60  \\
\midrule
BERT[0]      & ASR  & 46.61   & 45.83  \\
BERT[-3,3]       & ASR  & 64.15  & 65.73  \\
Audio$_{25,250}$+BERT[-3,3] & ASR  & 70.96  & 71.90  \\
\midrule
Audio$_{25,250}$+BERT[-3,3] & Mix  & 61.91  & 63.47  \\
\bottomrule
\end{tabular}
\caption{5-fold CV results with real ASR outputs. ``Ref" denotes reference transcripts provided in dataset. ``Mix" means that the system was trained on reference transcripts and tested on ASR output. ``Audio$_{25,250}$" denotes FBK$_{25}$+pitch+$\Delta$+FBK$_{250}$.}
\label{tab: ASR}
\end{table}
Since the Google API does not provide the word-to-frame alignments, the GloVe feature was not used in the experiments in this section. The results are shown in Table~\ref{tab: ASR}. The use of ASR leads to an accuracy decrease of $\sim$12.6\% and  $\sim$6.6\% for BERT[0] and BERT[-3,3], respectively. The decrease is reduced to $\sim$3.7\% when the audio features were included. Comparing the last two rows in Table~\ref{tab: ASR}, the system that is both trained and tested with ASR outputs performed better than the system trained on reference transcriptions but tested on the ASR outputs. This may reflect the fact that the system can learn to better account for the errorful ASR transcipts when they are also used in training.
%be due to the fact that training on ASR outputs can reduce the mismatch between the transcriptions used in training and test, by capturing some error patterns produced by the ASR. %For example, the system might have learnt not to rely too much on a single modality. 
A major advantage of using multimodal features is  that different modalities can augment or complement each other, especially when certain modalities are susceptible to noise.
%only BERT embeddings extracted from ASR transcriptions and audio features are used 
%\begin{figure}[htbp!] 
%\centering    
%\includegraphics[width=\linewidth]{fig/asr_new.PNG}
%\caption{Performance drop caused by using transcription generated by ASR instead of reference transcripts.}
%\label{fig: ASR}
%\end{figure}

%As shown in Figure~\ref{fig: ASR},
Further from the results shown in Table~\ref{tab: ASR}, 
the performance loss caused by using ASR transcriptions is smaller when cross-utterance information was added, which matches the expected findings discussed in Sec.~\ref{sec:tsbtab}. The addition of context can help in two ways. First, it provides more information, and second, it can partly compensate for the case that ASR system gives no valid output. %Splitting out the unrecognised utterances and testing only on those utterances, leads to the results shown in Table~\ref{tab: unreg}. Context mode produces much better results on unrecognised utterances as expected. The UA of BERT[0]-ASR is exactly equal to random guess of the 4-way classification. The WA of BERT[0]-ASR relies mostly on prior information only. It shows the information lost due to empty ASR outputs can be partly recovered by using the transcriptions of the context utterances, which makes the system more robust. %Incorporating context sentences provides more information and makes the system more robust.
Analysing the 9.7\% of utterances without valid ASR output and looking at AER results for only those utterances, the BERT[0] system gave 25.00\% UA, which is the same as a random guess. However, BERT[-3,3] produced a much better UA of 57.34\%, which shows the information lost due to ASR failures can be partly recovered with the context utterances, and hence makes the system more robust.

%\begin{table}[htbp!]
%\centering
%\begin{tabular}{ccc}
%\toprule
%Feature   & WA (\%) & UA (\%) \\
%\midrule
%BERT[0]-ASR & 31.18  & 25.00  \\
%BERT[-3,3]-ASR & 57.90  & 57.34 \\
%\bottomrule
%\end{tabular}
%\caption{5-fold CV averages on utterances that cannot be recognised by the Google ASR API. The BERT embeddings of the unrecognised utterances are the outputs of an empty sentence.}
%\label{tab: unreg}
%\end{table}
\vspace{-1mm}
\subsection{5-way classification results}
One of the main goals of this paper is to evaluate the performance of an emotion recognition system without making any unrealistic assumptions. Such unrealistic assumptions include assuming perfect ASR results, and that only a subset of emotion types will be seen in real applications. Papers that investigate performance on IEMOCAP normally only use a 4-way classification task based on the emotions: happy, sad, angry, neutral. However, in reality, people express many other emotions, such as fear, surprise, and disgust. In fact, IEMOCAP itself contains 10 different emotion labels, and when only 4 emotions are considered, nearly half of the utterances in the dataset are normally discarded. 
% IEMOCAP contains 7,532 utterances with ground truth labels. Previously only 5,531 of those utterances with labels from for 4-way classification (happy, sad, neutral, angry) were used for the training and testing of the the AER systems. 
% However, when considering a more realistic setup, 
% there are various other emotions that occur, such as ``frustration'', ``fear'', ``surprise'', and ``disgust'', which will be incorrectly classified by the 4-way system. 
Therefore, in this section we investigate an alternative 5-way classification setup which has an extra class ``others'' to represent all the other emotions that exist in IEMOCAP. 
The data used for  the ``others'' class includes utterances labelled as ``frustration", ``fear'', ``surprise'', ``disgust'', and ``other". Note that ``frustration'' accounts for 92.4\% of the data for the ``others'' category and is also the largest emotion group in IEMOCAP. 
In the 5-way classification setup, all of the 7,532 utterances with ground truth labels were used for training and testing. %The 5-fold CV AER classification performance of our 5-way system achieves 77.16\% WA and 77.09\% UA on the testing sets of the 4-way setup. Although the classification accuracy is expected to decrease when changing a 4-way classification into a 5-way classification, it is not obvious comparing to the results shown in Table~\ref{tab: liter sum}. 
The classification accuracy of the 5-way system on the previous four emotions (happy, sad, neutral, angry) is 77.67\% WA and 77.74\% UA, which is similar to the results of the 4-way system given in Table~\ref{tab: liter sum}. %However, when the 4-way system encounters emotions other than these four, it don't know how to classify them. 
However, the reverse would not be the case since the 4-way system cannot correctly classify examples from the other class and the overall classification accuracy of the 4-way system drops dramatically to 57.02\% WA and 62.72\% UA when tested on the 5-way data. The results show that when encountering emotions that don’t belong to the target four emotions, while the 4-way system will obviously result in errors, our 5-way system can classify these other emotion classes with more than 75\% accuracy.

The final 5-fold CV results for the 5-class system with ASR transcriptions are 
shown in Table \ref{tab: ASR-5way}. Our 5-way system, combined with real ASR output transcriptions, can serve as a prototype for a more realistic emotion classification system. %, which is an additional contribution of this paper. 

% , which can be viewed as the performance of a more realistic AER system.   

%Reference transcripts are then replaced by ASR transcriptions, 5cv results shown in Table~\ref{tab: ASR-5way}. 

% IEMOCAP in total contains 7532 utterances that have ground truth label. The previous 4-way classification setup only used less than 75\% of them. 

%, which indicates that the system proposed in this paper is able to cope with frustration, a relatively weak but important emotion.

\begin{table}[htbp!]
\centering
\begin{tabular}{cccc}
\toprule
Feature & Text & WA (\%) & UA (\%)\\
\midrule
Audio$_{25,250}$+BERT[-3,3]& Ref  & 73.34  & 74.44  \\
Audio$_{25,250}$+BERT[-3,3] & ASR  & 69.44  & 70.90  \\
\bottomrule
\end{tabular}
\caption{5-fold CV results of the 5-way system with the 5-way training and test data with different sources for the text modality.  }
\label{tab: ASR-5way}
\vspace{-5mm}
\end{table}

%audio+fbk250+bert7 ref

\iffalse
\begin{figure}[htbp!] 
\centering    
\includegraphics[width=0.9\linewidth]{fig/ground_truth.PNG}
\caption{Statistics of emotion label of utterances with ground truth in IEMOCOP. Excited merged with happy.}
\label{fig: label}
\end{figure}

\begin{table}[htbp!]
\centering
\begin{tabular}{ccc}
\hline
\#class & WA (\%) & UA (\%) \\
\hline
4 & 77.57  & 78.41 \\
5 & 77.16  & 77.09 \\
\hline
\end{tabular}
\caption{Average of 5-fold CV results on five-way classification when ``frustration" was included.}
\label{tab: 5-class}
\end{table}
\fi

\section{Conclusion}
\label{sec:conclusion}

In this paper, a two-branch structure fusing time synchronous and time asynchronous features has been proposed for multimodal emotion recognition. 
%The TSB captures the correlations between each word and its acoustic realisations by fusing the audio features with the corresponding text information at each time step, with the  TAB capturing meaning information in the speech transcriptions by integrating the sentence embeddings from a number of context utterances. 
The TSB captures the correlations between each word and its acoustic realisations at each time step, while the  TAB integrates the sentence embeddings from context utterances. 
The two-branch system achieves state-of-the-art 4-way classification accuracy of 77.76\% WA and 78.30\% UA for 10-fold CV on IEMOCAP. %By investigating more realistic settings such as the use of erroneous transcriptions produced by ASR and the alternative 5-way classification setup, we demonstrate the performance of a realistic AER system.
By investigating the use of transcriptions produced by an ASR system and the alternative 5-way classification setup with an extra class representing all the other emotions, the performance of a more realistic AER system has also been demonstrated.

%The TSB fuses two modalities at frame level which utilizes the alignment information to better model the correlation of word and its acoustic realization. Cross-utterance information incorporated by TAB improves performance as well as the robustness of the system. This structure works well in five-way classification task which handles more realistic cases. Apart from reference transcripts, transcriptions generated by ASR has also been investigated. The high WER shows that recognizing emotional speech is still a difficult task for ASR. 

\vfill\pagebreak

% -------------------------------------------------------------------------
%\section{References}
%\vspace{-0.5em}
%\begingroup
\bibliographystyle{IEEEbib}
\bibliography{strings,refs}
%\endgroup

\end{document}